\documentclass[letterpaper]{article} 
\usepackage[preprint]{aaai2027}  
\usepackage[hyphens]{url}  
\usepackage{graphicx} 
\usepackage{natbib}  
\usepackage{caption} 
\usepackage{algorithm}
\usepackage{algorithmic}
\usepackage{amsmath}    
\usepackage{multirow}   
\usepackage{booktabs}   
\usepackage{graphicx}  
\usepackage{tabularx}

\usepackage{newfloat}
\usepackage{listings}
\DeclareCaptionStyle{ruled}{labelfont=normalfont,labelsep=colon,strut=off} 
\floatstyle{ruled}
\newfloat{listing}{tb}{lst}{}
\floatname{listing}{Listing}

\usepackage{booktabs}

\title{Exploiting Intrinsic Duality for Multi-Hop Question Generation
}
\author{
    Maodong Li,
    Xinyue Kang, Yuanchen Shi, Fang Kong
}
\affiliations{
School of Computer Science and Technology, Soochow University, China \\
Jiangsu Key Lab of Language Computing, Suzhou 215123, China \\
\texttt{alimaodong@gmail.com}, \texttt{\{20244227013@stu,20227927002@stu,kongfang@\}suda.edu.cn}
}

\begin{document}

\maketitle

\begin{abstract}
Multi-hop question generation (MQG) aims to generate questions from multiple given documents and target answers, whereas question answering (QA) focuses on deriving answers from documents given specific questions. Although MQG and QA are inherently dual tasks, most existing MQG studies largely overlook this intrinsic duality. To address this limitation, we propose QQ, a novel framework that exploits the duality between \textbf{Q}uestion and answer for multi-hop \textbf{Q}uestion generation. Specifically, QQ employs a unified architecture functioning simultaneously as both an MQG and a QA model to fully leverage their interdependence. Our framework is driven by two key mechanisms: (i) enforcing bidirectional alignment constraints to ensure strict mutual correspondence between the questions generated by the MQG model and the answers produced by the QA model; and (ii) applying contrastive learning to pull paired question–answer representations closer while pushing unpaired ones apart, thereby reinforcing this correspondence. Extensive automatic and human evaluations on the HotpotQA and MuSiQue datasets demonstrate that the QQ framework significantly improves the quality of generated multi-hop questions.

\end{abstract}

\section{Introduction}
Question generation (QG) is the task of generating a question given a single document and a target answer. As a fundamental task in natural language processing, it benefits broad downstream applications such as automatic question answering and machine reading comprehension \cite{liang-etal-2023-prompting, ijcai2024p889, li-etal-2025-multi-hop}. In contrast to conventional QG, multi-hop question generation (MQG) aims to generate questions that span multiple documents and target answers, rendering it more representative of complex real-world scenarios \cite{chen2023toward,DBLP:conf/coling/Hwang0L24,kim-etal-2024-non}.

\begin{figure}[!t]
	\centering
	\includegraphics[width=\columnwidth]{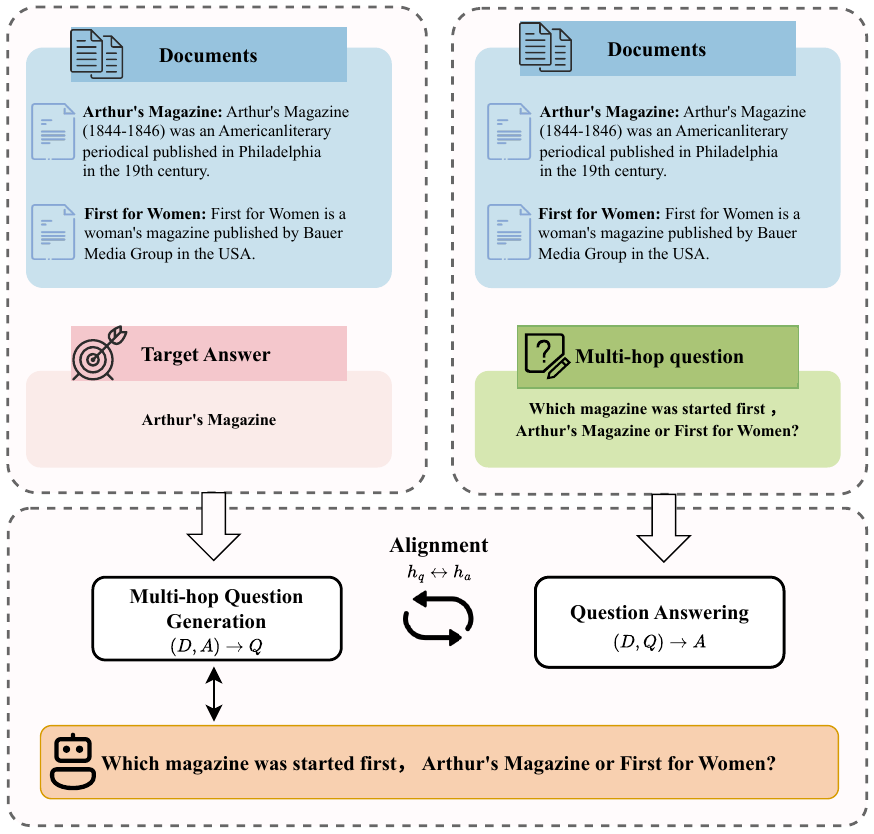}
	\caption{
		An illustration of a multi-hop question generated via alignment. During training, we enforce bidirectional alignment between MQG-generated questions and QA-generated answers.
	}
	\label{sample1}
\end{figure}

Generally, the performance of MQG depends on how effectively a model leverages both answer and document information, and maximizing the utilization of this information can improve the quality of the generated multi-hop questions. \citet{su-etal-2020-multi,su2022qa4qg} additionally employ a question answering (QA) model to utilize answer information, but only as part of the input, without deeply integrating it into the question generation process. Some advanced models focus on key sentences or keywords within documents to extract document information \cite{ding-etal-2024-sgcm, li-etal-2025-multi-hop}. However, while these representative methods aim to enhance the use of answer or document information, they fail to exploit the intrinsic duality between MQG and QA, which necessitates modeling document, question, and answer information simultaneously. Motivated by this observation, we treat MQG and QA as mutually dependent processes and explore their bidirectional synergy to enhance MQG performance, as illustrated in Figure~\ref{sample1}. For example, given input documents and a target answer (\textit{Arthur's Magazine}), the model generates a complex multi-hop question (\textit{Which magazine ...}). Conversely, the model leverages the documents and the provided question to predict the target answer. The core principle lies in enforcing mutual alignment: during multi-hop question generation, the generated questions should align with the corresponding answers from the QA model, and simultaneously, during question answering, the predicted answers should align with the corresponding questions from the MQG model.

To this end, we propose QQ, a framework that exploits the intrinsic duality between \textbf{Q}uestion and answer for multi-hop \textbf{Q}uestion generation. QQ employs a unified architecture that simultaneously serves as both an MQG and a QA model, fully exploiting their intrinsic duality to enhance MQG performance. We achieve this alignment by modeling the mutual predictability between MQG-generated questions and QA-generated answers. Specifically, during training, we model the predictability between MQG-generated question states mapped into a predictive space and the corresponding QA-generated answer states, as well as the reciprocal predictability from mapped QA-generated answer states to the corresponding MQG-generated question states. This mechanism enforces bidirectional alignment constraints to synchronize the latent representations of generated questions and their corresponding answers. To further strengthen this alignment, we introduce contrastive learning, mapping the question and answer states into a contrastive space such that paired question–answer representations are pulled closer while unpaired ones are pushed farther apart. During inference, only the MQG model is employed, taking the input documents and target answers to generate multi-hop questions. 

To verify the effectiveness of the QQ framework, we conduct systematic automatic and human evaluations on two widely used datasets, HotpotQA and MuSiQue. Experimental results demonstrate that the QQ framework effectively addresses both multi-hop question generation and question answering tasks, and significantly improves the quality of the generated multi-hop questions. Our contributions can be summarized as follows:

\begin{itemize}
	\item We advance MQG by explicitly modeling its alignment with the QA process during generation, leveraging the intrinsic duality between the two tasks to significantly enhance question generation performance.
	\item We propose an alignment strategy implemented by modeling the mutual predictability between MQG-generated questions and QA-generated answers, which is further augmented with contrastive learning to effectively distinguish paired from unpaired latent representations.
\end{itemize}

\section{Related Work}
Question generation can be broadly categorized into single-hop (SQG) and multi-hop (MQG) paradigms. Accordingly, we review the relevant research in both areas.

\subsection{Single-hop Question Generation}
Single-hop question generation is the task of generating a question given a single document and a target answer. Early approaches to SQG were predominantly rule-based \cite{heilman-smith-2010-good, chali-hasan-2012-towards, mazidi-nielsen-2014-linguistic}, but their performance was constrained by limited flexibility and generalizability. As the field advanced, attention-based sequence-to-sequence architectures gained prominence \cite{du-etal-2017-learning,10.1007/978-3-319-73618-1_56, sun-etal-2018-answer,song-etal-2018-leveraging}. Subsequent studies incorporated feature-rich encoders to capture broader contextual information \cite{10.1007/978-3-319-73618-1_56,sun-etal-2018-answer} and introduced answer-aware mechanisms to extract more precise answer signals \cite{sun-etal-2018-answer,song-etal-2018-leveraging}. To address the challenges posed by long passages, a maxout pointer mechanism was developed \cite{zhao-etal-2018-paragraph}. Additionally, reinforcement learning has been applied within graph-to-sequence models for question generation \cite{DBLP:conf/iclr/0022WZ20, chen2023toward}. These studies primarily address SQG, with the SQuAD dataset \cite{rajpurkar-etal-2016-squad} serving as the popular benchmark.

\subsection{Multi-hop Question Generation}
In contrast to conventional SQG, multi-hop question generation is significantly more demanding and has attracted increasing attention in recent years \cite{su-etal-2020-multi,fei-etal-2021-iterative,fei-etal-2022-cqg,chen2023toward,ding-etal-2024-sgcm,li-etal-2025-multi-hop}, constituting the primary focus of our work. Early studies applied graph neural networks (GNNs) to multi-hop reasoning to facilitate question generation \cite{su-etal-2020-multi,pan-etal-2020-semantic}. However, with the rapid advancement of transformer-based architectures, large pretrained language models have demonstrated superior performance over GNNs in this task \cite{vaswani2017attention,su2022qa4qg}. To further enhance the capabilities of pretrained language models in MQG, numerous studies have explored auxiliary tasks, proposing methods such as order-agnostic learning and sequential rewriting to improve generation quality \cite{murakhovska-etal-2022-mixqg,kim-etal-2024-non,DBLP:conf/coling/Hwang0L24}. Generally, these auxiliary strategies focus on two main directions: (i) leveraging answer information, typically by incorporating an auxiliary QA model as an external input \cite{su-etal-2020-multi,su2022qa4qg}; and (ii) capturing contextual information from documents more effectively via intermediate extraction models \cite{ding-etal-2024-sgcm,DBLP:conf/emnlp/XiaG0YHLN23,li-etal-2025-multi-hop}.

However, existing methods primarily utilize answer or document information in isolation, leaving the intrinsic duality between MQG and QA largely under-explored. This duality requires jointly modeling questions and answers. To address this gap, we leverage bidirectional alignment between the two tasks as the core mechanism for improving MQG performance.
\begin{figure}[!t]
	\centering
	\includegraphics[width=\columnwidth]{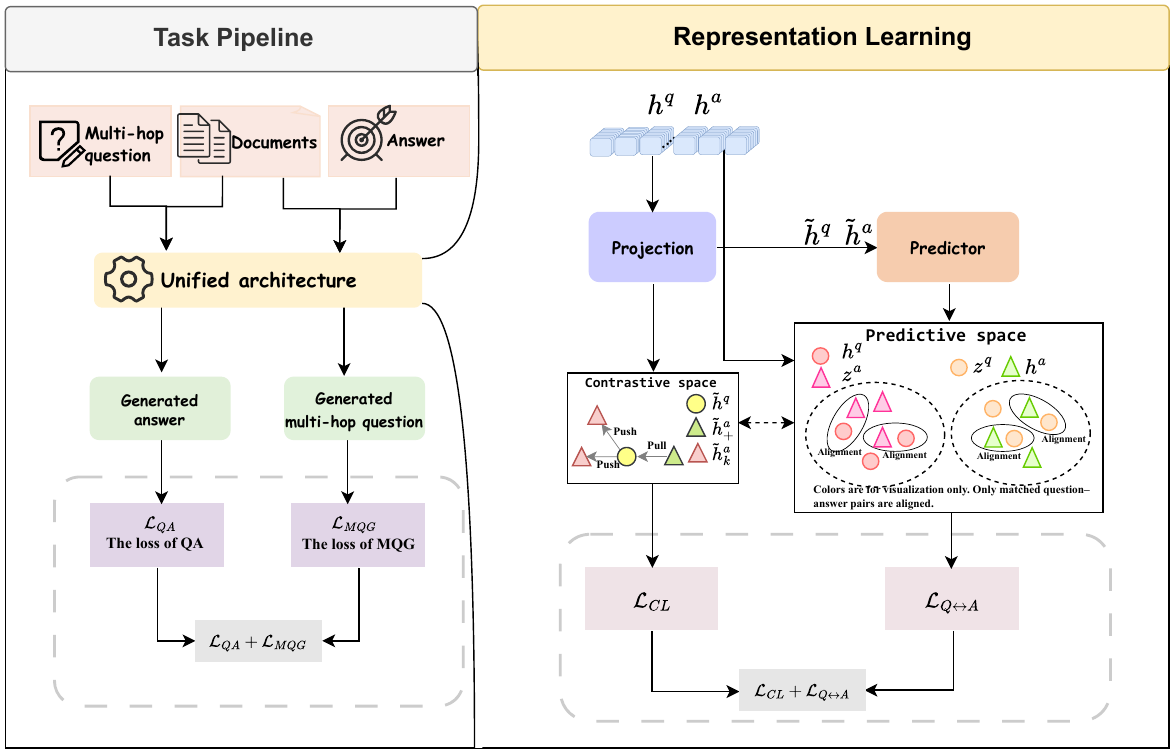}
	\caption{
		Our QQ framework.
	}
	\label{fig01}
\end{figure}

\section{Methodology}
\subsection{Task Formulation}
Let $\mathcal{U} = \{(\mathcal{D}^{(i)}, \mathcal{A}^{(i)}, \mathcal{Q}^{(i)})\}_{i=1}^{N}$ be an MQG dataset, where $\mathcal{D}^{(i)}$ denotes the context documents, $\mathcal{A}^{(i)}$ represents the target answer, $\mathcal{Q}^{(i)}$ is the corresponding multi-hop question, and $N$ is the total number of samples. For the MQG task, the objective is to generate a multi-hop question $\hat{\mathcal{Q}}^{(i)}$ given $\mathcal{D}^{(i)}$ and $\mathcal{A}^{(i)}$. For the QA task, the goal is to generate an answer $\hat{\mathcal{A}}^{(i)}$ given $\mathcal{D}^{(i)}$ and $\mathcal{Q}^{(i)}$.

\subsection{Framework Overview}
In this section, we present the proposed QQ framework in detail, with its overall architecture illustrated in Figure~\ref{fig01}. The QQ framework employs a unified architecture that functions simultaneously as both an MQG and a QA model: it generates a multi-hop question given documents and a target answer, and conversely generates an answer given documents and a provided question.

During the MQG process, question states are mapped into a predictive space via the projection and predictor modules to evaluate their predictability with respect to the corresponding QA-generated answer states. Conversely, during the QA process, answer states are mapped into the shared predictive space using identical modules to compute their predictability against the MQG-generated question states. This mechanism enforces bidirectional alignment constraints between MQG and QA to synchronize the latent representations of their respective outputs. To further strengthen this alignment, we integrate contrastive learning, wherein question states from the MQG model and answer states from the QA model are mapped into a contrastive space via the projection module, effectively pulling paired question–answer representations closer together while pushing unpaired ones farther apart.

\subsection{MQG and QA Pipeline}
For the MQG task, we represent the output multi-hop question $\mathcal{Y}^q$ at the token level as $\mathcal{Y}^q = (y^q_1, y^q_2, \dots, y^q_{T_q})$ of length $T_q$, conditioned on the input text sequence $x_{MQG} = (\mathcal{D}, \mathcal{A})$. Similarly, for the QA task, the output answer $\mathcal{Y}^a$ is represented as $\mathcal{Y}^a = (y^a_1, y^a_2, \dots, y^a_{T_a})$ of length $T_a$, conditioned on $x_{QA} = (\mathcal{D}, \mathcal{Q})$. The conditional distributions for MQG and QA are formulated as follows:
\begin{equation}
	p^q_{\theta}(y^q_t|y^q_{<t},x_{MQG}) = \mathrm{softmax}(Wh^q_t + b)
\end{equation}
\begin{equation}
	p^a_{\theta}(y^a_t|y^a_{<t},x_{QA}) = \mathrm{softmax}(Wh^a_t + b)
\end{equation}
where $W$ and $b$ are trainable parameters, and $\theta$ denotes the MQG and QA model parameters. Both the MQG and QA models are optimized by minimizing the negative log-likelihood over $N$ observations, denoted as $\{(x_{MQG}^{(i)}, y^{q(i)})\}_{i=1}^N$ and $\{(x_{QA}^{(i)}, y^{a(i)})\}_{i=1}^N$ respectively, which are formally expressed as:
\begin{equation}
	\mathcal{L}_{MQG} = -\sum^N_{i=1}\sum^{T_q}_{t=1} \log p^q_{\theta}(y^{q(i)}_t|y^{q(i)}_{<t},x_{MQG}^{(i)})
\end{equation}

\begin{equation}
	\mathcal{L}_{QA} = -\sum^N_{i=1}\sum^{T_a}_{t=1} \log p^a_{\theta}(y^{a(i)}_t|y^{a(i)}_{<t},x_{QA}^{(i)})
\end{equation}

\begin{table*}[!t]
	\centering
	\small
	\begin{tabular}{l ccc @{\hspace{2em}} ccccc}
		\toprule
		\multirow{2}{*}{Split} & \multicolumn{3}{c}{HotpotQA} & \multicolumn{5}{c}{MuSiQue} \\
		\cmidrule(r){2-4} \cmidrule(l){5-9}
		& \#Sample & Len. (Full) & Len. (SF) & \#Total & \#2-hop & \#3-hop & \#4-hop & Len. \\
		\midrule
		Train & 89,947 & 204.4 & 96.1 & 19,938 & 14,376 & 4,387 & 1,175 & 362.3 \\
		Dev   & 500    & 199.8 & 95.5 & 1,150  & 600    & 350   & 200   & 368.8 \\
		Test  & 7,405  & 205.8 & 98.5 & 1,267  & 652    & 410   & 205   & 369.5 \\
		\bottomrule
	\end{tabular}
	\caption{Dataset statistics. Len. denotes the average context length. The average lengths for the 2-, 3-, and 4-hop instances in MuSiQue are 238.4, 385.9, and 476.3, respectively.}
	\label{table_datasets_combined}
\end{table*}

\subsection{Bidirectional Alignment and Contrastive Objectives}
We achieve this bidirectional alignment by modeling the mutual predictability between MQG-generated questions and QA-generated answers. Specifically, within a shared predictive space, QA-generated answers can be predicted from their corresponding MQG-generated questions, and vice versa. This predictability is realized through an asymmetric similarity mechanism, which maps the latent representations of both task outputs into the predictive space to encourage cross-directional similarity. Consequently, the alignment losses, $\mathcal{L}_{Q \to A}$ and $\mathcal{L}_{A \to Q}$, are formulated as follows:
\begin{equation}
	\mathcal{L}_{Q \to A} = -\log \left( \frac{\exp(\mathrm{sim}(z^q,h^a_{+})/ \tau_1)}{\sum_{h^a_k \in V^a}\exp(\mathrm{sim}(z^q,h^a_k)/ \tau_1)} \right)
\end{equation}
\begin{equation}
	\mathcal{L}_{A \to Q} = -\log \left( \frac{\exp(\mathrm{sim}(z^a,h^q_{+})/ \tau_1)}{\sum_{h^q_k \in V^q}\exp(\mathrm{sim}(z^a,h^q_k)/ \tau_1)} \right)
\end{equation}
\begin{equation}
	z^q = \mathrm{pred}(\mathrm{proj}(h^q)), \quad z^a = \mathrm{pred}(\mathrm{proj}(h^a))
\end{equation}
where $\mathrm{pred}(\cdot)$ and $\mathrm{proj}(\cdot)$ denote the predictor and projection modules shown in Figure~\ref{fig01}, both implemented as a two-layer multilayer perceptron with a ReLU activation function. $z^q$ and $z^a$ represent the question and answer states in the predictive space, respectively, while $h^a_{+}$ and $h^q_{+}$ denote the corresponding predictable answer and question states. $V^a$ and $V^q$ refer to the sets of unmatched answer and question states corresponding to $z^q$ and $z^a$, respectively, and $\mathrm{sim}(\cdot)$ denotes the cosine similarity function. To stabilize the training process, we introduce a coefficient $\tau_1$. Structurally, this formulation resembles standard contrastive learning, differing primarily in the definitions of positive and negative samples. However, unlike standard contrastive loss, the proposed alignment loss does not impose a symmetry constraint; instead, it captures an implicit inference relationship between MQG-generated questions and QA-generated answers \cite{grill2020bootstrap}. Overall, $\mathcal{L}_{Q \to A}$ and $\mathcal{L}_{A \to Q}$ mutually align MQG-generated questions and their corresponding QA-generated answers, synchronizing their latent representations in accordance with our motivation. Furthermore, we define the final bidirectional alignment loss $\mathcal{L}_{Q \leftrightarrow A}$ as the average of $\mathcal{L}_{Q \to A}$ and $\mathcal{L}_{A \to Q}$ (hereafter referred to as the alignment loss):
\begin{equation}
	\mathcal{L}_{Q \leftrightarrow A} = \frac{1}{2}(\mathcal{L}_{Q \to A} + \mathcal{L}_{A \to Q})
\end{equation}
We observe that introducing an additional contrastive learning space further enhances the mutual alignment between MQG-generated questions and their corresponding QA-generated answers, acting in a mutually reinforcing manner with $\mathcal{L}_{Q \leftrightarrow A}$, as illustrated in Figure~\ref{fig01}. The contrastive learning loss, denoted as $\mathcal{L}_{CL}$, is formulated as follows:
\begin{equation}
	\mathcal{L}_{CL} = -\log \left( \frac{\exp(\mathrm{sim}(\tilde{h}^q,\tilde{h}^a_{+})/ \tau_2)}{\sum_{\tilde{h}^a_k \in S^a}\exp(\mathrm{sim}(\tilde{h}^q,\tilde{h}^a_k)/ \tau_2)} \right)
\end{equation}
\begin{equation}
	\tilde{h}^q = \mathrm{proj}(h^q), \quad \tilde{h}^a = \mathrm{proj}(h^a)
\end{equation}
where $\tilde{h}^q$ and $\tilde{h}^a$ denote the question and answer states in the contrastive space, respectively. $\tilde{h}^a_{+}$ represents the corresponding positive answer state for $\tilde{h}^q$, $\tau_2$ is a temperature parameter, and $S^a$ denotes the set of negative answer states unpaired with $\tilde{h}^q$. Finally, the overall training loss $\mathcal{L}$ of the QQ framework is defined as:
\begin{equation}
	\mathcal{L} = \mathcal{L}_{MQG} + \mathcal{L}_{QA} + \mathcal{L}_{Q \leftrightarrow A} +  \mathcal{L}_{CL}
\end{equation}
With the aforementioned training objectives, the proposed framework is fully implemented.

\begin{table*}[!t]
	\small
	\centering
	
	\centerline{\textbf{(a) Results on HotpotQA Dataset}}
	\vspace{1mm}
	\begin{tabular}{lllllll}
		\toprule
		\multirow{2}{*}{Model} & \multicolumn{3}{c}{SF setting} & \multicolumn{3}{c}{Full setting} \\ \cmidrule(l){2-7} 
		& BLEU-4 & METEOR & ROUGE-L & BLEU-4 & METEOR & ROUGE-L \\ \midrule
		MulQG & - & - & - & 15.20 & 20.51 & 35.30 \\
		CQG & 25.09 & 27.45 & 41.83 & 21.46 & 24.97 & 39.61 \\
		MixQG & 25.45 & 26.36 & 43.21 & 22.13 & 23.78 & 41.21 \\
		QA4QG$_{\text{large}}$ & 25.70 & 27.44 & 46.48 & 21.21 & 25.53 & 42.44 \\
		E2EQR & 21.73 & 27.47 & 41.34 & - & - & - \\
		SGCM & 26.16 & 28.51 & 44.06 & 22.61 & 26.04 & 40.61 \\
		DPKG$_{\text{hard}}$ & 26.80 & 27.87 & 46.50 & 22.74 & 24.90 & 43.29 \\
		DPKG$_{\text{soft}}$ & 26.19 & 28.51 & 46.36 & 23.33 & 25.21 & 43.18 \\ \midrule
		GPT$_{\text{(117M)}}$ & 18.27 & 21.57 & 39.89 & 15.33 & 19.06 & 37.09 \\ 
		\quad w/ QQ & \textbf{19.64}$_{\uparrow1.37}$ & \textbf{22.77}$_{\uparrow1.20}$ & \underline{40.33}$_{\uparrow0.44}$ & \textbf{16.36}$_{\uparrow1.03}$ & \textbf{20.13}$_{\uparrow1.07}$ & \underline{37.86}$_{\uparrow0.77}$ \\ \cmidrule(l){1-7}
		GPT$_{\text{(345M)}}$ & 19.31 & 22.59 & 40.82 & 16.60 & 20.47 & 38.75 \\  
		\quad w/ QQ & \textbf{20.77}$_{\uparrow1.46}$ & \textbf{23.86}$_{\uparrow1.27}$ & \textbf{41.91}$_{\uparrow1.09}$ & \underline{17.47}$_{\uparrow0.87}$ & \underline{21.22}$_{\uparrow0.75}$ & \underline{39.17}$_{\uparrow0.42}$ \\ \cmidrule(l){1-7}
		GPT$_{\text{(762M)}}$ & 19.33 & 22.84 & 41.15 & 16.37 & 20.13 & 38.87 \\ 
		\quad w/ QQ & \textbf{21.15}$_{\uparrow1.82}$ & \textbf{24.03}$_{\uparrow1.19}$ & \underline{41.99}$_{\uparrow0.84}$ & \underline{17.30}$_{\uparrow0.93}$ & \underline{20.81}$_{\uparrow0.68}$ & \underline{39.24}$_{\uparrow0.37}$ \\ \midrule
		LLaMA$_{\text{(1B)}}$ & 22.58 & 24.81 & 44.23 & 19.20 & 22.12 & 41.44 \\ 
		\quad w/ QQ & \underline{23.12}$_{\uparrow0.54}$ & \underline{25.23}$_{\uparrow0.42}$ & \underline{44.49}$_{\uparrow0.26}$ & \underline{19.67}$_{\uparrow0.47}$ & \underline{22.64}$_{\uparrow0.52}$ & \underline{41.61}$_{\uparrow0.17}$ \\ \cmidrule(l){1-7}
		LLaMA$_{\text{(3B)}}$ & 23.49 & 25.57 & 44.98 & 19.73 & 22.29 & 42.48 \\ 
		\quad w/ QQ & \textbf{24.63}$_{\uparrow1.14}$ & \underline{26.29}$_{\uparrow0.72}$ & \textbf{46.13}$_{\uparrow1.15}$ & \textbf{21.58}$_{\uparrow1.85}$ & \textbf{24.18}$_{\uparrow1.89}$ & \underline{42.71}$_{\uparrow0.23}$ \\ \cmidrule(l){1-7}
		LLaMA$_{\text{(8B)}}$ & 25.90 & 27.30 & 47.12 & 21.18 & 23.44 & 43.66 \\ 
		\quad w/ QQ & \underline{26.83}$_{\uparrow0.93}$ & \underline{27.96}$_{\uparrow0.66}$ & 46.84$_{\downarrow0.28}$ & \textbf{23.71}$_{\uparrow2.53}$ & \textbf{25.41}$_{\uparrow1.97}$ & \underline{44.20}$_{\uparrow0.54}$ \\ \bottomrule
	\end{tabular} 
	
	\vspace{6mm} 
	
	\centerline{\textbf{(b) Results on MuSiQue Dataset}}
	\vspace{1mm}
	\begin{tabular}{llll|lll|lll}
		\toprule
		\multirow{2}{*}{Model} & \multicolumn{3}{c|}{2-hop} & \multicolumn{3}{c|}{3-hop} & \multicolumn{3}{c}{4-hop} \\ \cmidrule(l){2-10} 
		& BLEU4 & METEOR & ROUGE-L & BLEU4 & METEOR & \multicolumn{1}{l|}{ROUGE-L} & BLEU4 & METEOR & ROUGE-L \\ \midrule
		DP-Graph & 5.14 & 10.80 & 28.69 & 4.33 & 10.37 & \multicolumn{1}{l|}{28.33} & 4.47 & 9.87 & 28.01 \\
		MulQG & 9.56 & 15.41 & 37.18 & 9.23 & 14.35 & \multicolumn{1}{l|}{35.66} & 7.13 & 12.43 & 31.88 \\
		CQG & 9.64 & 16.20 & 33.98 & 7.79 & 13.77 & \multicolumn{1}{l|}{31.12} & 5.14 & 11.93 & 26.48 \\
		E2EQR & 20.33 & 25.64 & 44.01 & 17.02 & 22.33 & \multicolumn{1}{l|}{40.04} & 15.34 & 19.78 & 36.98 \\ \midrule
		LLaMA$_{\text{(1B)}}$ & 16.30 & 22.08 & 42.29 & 15.78 & 20.17 & \multicolumn{1}{l|}{42.14} & 9.34 & 15.70 & 34.83 \\
		\quad w/ QQ & 15.54$_{\downarrow0.76}$ & \textbf{23.14}$_{\uparrow1.06}$ & 41.67$_{\downarrow0.62}$ & \textbf{17.16}$_{\uparrow1.38}$ & \underline{20.87}$_{\uparrow0.70}$ & \multicolumn{1}{l|}{\underline{42.50}$_{\uparrow0.36}$} & \textbf{11.80}$_{\uparrow2.46}$ & \textbf{16.70}$_{\uparrow1.00}$ & \textbf{35.96}$_{\uparrow1.13}$ \\
		LLaMA$_{\text{(3B)}}$ & 17.03 & 24.94 & 44.05 & 19.08 & 22.86 & \multicolumn{1}{l|}{44.41} & 13.19 & 18.38 & 35.29 \\
		\quad w/ QQ & \textbf{18.81}$_{\uparrow1.78}$ & \underline{25.75}$_{\uparrow0.81}$ & \textbf{45.55}$_{\uparrow1.50}$ & \underline{19.33}$_{\uparrow0.25}$ & 22.31$_{\downarrow0.55}$ & \multicolumn{1}{l|}{44.37$_{\downarrow0.04}$} & 12.67$_{\downarrow0.52}$ & 17.75$_{\downarrow0.63}$ & \textbf{36.77}$_{\uparrow1.48}$ \\
		LLaMA$_{\text{(8B)}}$ & 19.34 & 26.38 & 45.23 & 19.72 & 23.33 & \multicolumn{1}{l|}{43.50} & 15.77 & 19.22 & 39.12 \\
		\quad w/ QQ & 19.22$_{\downarrow0.12}$ & \underline{27.10}$_{\uparrow0.72}$ & \textbf{46.98}$_{\uparrow1.75}$ & \underline{20.13}$_{\uparrow0.41}$ & 23.31$_{\downarrow0.02}$ & \multicolumn{1}{l|}{\textbf{44.54}$_{\uparrow1.04}$} & 15.04$_{\downarrow0.73}$ & \underline{19.96}$_{\uparrow0.74}$ & \underline{39.72}$_{\uparrow0.60}$ \\ \bottomrule
	\end{tabular}
	
	\caption{
		Overall performance comparison on the HotpotQA and MuSiQue datasets. Improvements of less than 1.0 over the original backbone are \underline{underlined}, while improvements greater than 1.0 are highlighted in \textbf{bold}.
	}
	\label{mainresult_merged}
\end{table*}

\begin{table*}[!t]
	\small
	\centering
	\begin{tabular}{llllllllll}
		\toprule
		\multicolumn{2}{l}{\multirow{2}{*}{Model}} & \multicolumn{4}{c}{SF setting} & \multicolumn{4}{c}{Full setting} \\ \cmidrule(l){3-10} 
		\multicolumn{2}{c}{} & BLEU-4 & METEOR & ROUGE-L & BERTScore & BLEU-4 & METEOR & ROUGE-L & BERTScore \\ \midrule
		\multirow{4}{*}{GPT$_{\text{(117M)}}$} & w/ QQ & 19.64$_{\uparrow1.37}$ & 22.77$_{\uparrow1.20}$ & 40.33$_{\uparrow0.44}$ & 48.39$_{\uparrow0.68}$ & 16.36$_{\uparrow1.03}$ & 20.13$_{\uparrow1.07}$ & 37.86$_{\uparrow0.77}$ & 45.62$_{\uparrow0.78}$ \\
		& w/ $\mathcal{L}_{Q \leftrightarrow A}$ & 19.26$_{\uparrow0.99}$ & 22.38$_{\uparrow0.81}$ & \textbf{40.67}$_{\uparrow0.78}$ & \textbf{48.78}$_{\uparrow1.07}$ & \textbf{16.25}$_{\uparrow0.92}$ & \textbf{20.02}$_{\uparrow0.96}$ & 37.79$_{\uparrow0.70}$ & \textbf{45.52}$_{\uparrow0.68}$ \\
		& w/ $\mathcal{L}_{CL}$ & \underline{19.57}$_{\uparrow1.30}$ & \underline{22.73}$_{\uparrow1.16}$ & 40.62$_{\uparrow0.73}$ & 48.68$_{\uparrow0.97}$ & 16.02$_{\uparrow0.69}$ & 19.68$_{\uparrow0.62}$ & \underline{37.82}$_{\uparrow0.73}$ & 45.51$_{\uparrow0.67}$ \\
		& vanilla & 18.27 & 21.57 & 39.89 & 47.71 & 15.33 & 19.06 & 37.09 & 44.84 \\ \cmidrule(l){2-10} 
		\multirow{4}{*}{GPT$_{\text{(345M)}}$} & w/ QQ & 20.77$_{\uparrow1.46}$ & 23.86$_{\uparrow1.27}$ & 41.91$_{\uparrow1.09}$ & 50.18$_{\uparrow1.12}$ & 17.47$_{\uparrow0.87}$ & 21.22$_{\uparrow0.75}$ & 39.17$_{\uparrow0.42}$ & 47.23$_{\uparrow0.42}$ \\
		& w/ $\mathcal{L}_{Q \leftrightarrow A}$ & 20.27$_{\uparrow0.96}$ & \textbf{23.69}$_{\uparrow1.10}$ & 41.24$_{\uparrow0.42}$ & 49.70$_{\uparrow0.64}$ & \textbf{17.10}$_{\uparrow0.50}$ & \textbf{20.72}$_{\uparrow0.25}$ & \textbf{39.05}$_{\uparrow0.30}$ & \textbf{47.04}$_{\uparrow0.23}$ \\
		& w/ $\mathcal{L}_{CL}$ & \underline{20.42}$_{\uparrow1.11}$ & \underline{23.69}$_{\uparrow1.10}$ & \underline{41.69}$_{\uparrow0.87}$ & \underline{49.94}$_{\uparrow0.88}$ & 16.64$_{\uparrow0.04}$ & 20.66$_{\uparrow0.19}$ & 38.38$_{\downarrow0.37}$ & 46.43$_{\downarrow0.38}$ \\
		& vanilla & 19.31 & 22.59 & 40.82 & 49.06 & 16.60 & 20.47 & 38.75 & 46.81 \\ \cmidrule(l){2-10} 
		\multirow{4}{*}{GPT$_{\text{(762M)}}$} & w/ QQ & 21.15$_{\uparrow1.82}$ & 24.03$_{\uparrow1.19}$ & 41.99$_{\uparrow0.84}$ & 50.13$_{\uparrow0.46}$ & 17.30$_{\uparrow0.93}$ & 20.81$_{\uparrow0.68}$ & 39.24$_{\uparrow0.37}$ & 47.08$_{\uparrow0.23}$ \\
		& w/ $\mathcal{L}_{Q \leftrightarrow A}$ & \textbf{21.06}$_{\uparrow1.73}$ & \textbf{23.98}$_{\uparrow1.14}$ & \textbf{42.36}$_{\uparrow1.21}$ & \textbf{50.62}$_{\uparrow0.95}$ & 17.22$_{\uparrow0.85}$ & \textbf{21.03}$_{\uparrow0.90}$ & \textbf{39.01}$_{\uparrow0.14}$ & \textbf{47.09}$_{\uparrow0.24}$ \\
		& w/ $\mathcal{L}_{CL}$ & 20.27$_{\uparrow0.94}$ & 23.41$_{\uparrow0.57}$ & 41.91$_{\uparrow0.76}$ & 50.24$_{\uparrow0.57}$ & \underline{17.24}$_{\uparrow0.87}$ & 20.85$_{\uparrow0.72}$ & 38.99$_{\uparrow0.12}$ & 46.94$_{\uparrow0.09}$ \\
		& vanilla & 19.33 & 22.84 & 41.15 & 49.67 & 16.37 & 20.13 & 38.87 & 46.85 \\ \bottomrule
	\end{tabular} 
	\caption{
		Ablation study of the QQ framework using the GPT series on the HotpotQA dataset. \textbf{Bold} indicates $\mathcal{L}_{Q \leftrightarrow A}$ dominance, while \underline{underlined} indicates $\mathcal{L}_{CL}$ dominance.
	}
	\label{ablation_study}
\end{table*}

\section{Experiments}
We discuss the main results in the main text to support our core findings, while providing comprehensive experiments and detailed analyses in the supplementary material.
\subsection{Experimental Setting}
\paragraph{Dataset} 
We conduct experiments on two widely used MQG datasets: \textbf{HotpotQA}~\cite{yang-etal-2018-hotpotqa} and \textbf{MuSiQue}~\cite{trivedi-etal-2022-musique}. For the HotpotQA dataset, we evaluate QQ under two settings: the SF setting, which provides only sentences containing supporting facts, and the Full setting, which provides longer and more complex contextual documents. The MuSiQue dataset consists of 2-hop, 3-hop, and 4-hop reasoning questions, further increasing the task difficulty. Detailed statistics for both datasets are summarized in Table~\ref{table_datasets_combined}.

\paragraph{Baselines and Evaluation Metrics}
We compare the proposed QQ framework against several recent advanced baselines: \textbf{MulQG} \cite{su-etal-2020-multi}, \textbf{CQG} \cite{fei-etal-2022-cqg}, \textbf{MixQG} \cite{murakhovska-etal-2022-mixqg}, \textbf{QA4QG} \cite{su2022qa4qg}, \textbf{E2EQR} \cite{DBLP:conf/coling/Hwang0L24}, \textbf{SGCM} \cite{ding-etal-2024-sgcm}, and \textbf{DPKG}$_{\text{hard/soft}}$ \cite{li-etal-2025-multi-hop}. To validate the generalizability of our framework across different architectures and model capacities, we employ two backbone families at varying scales: \textbf{GPT} \cite{radford2019language} (117M, 345M, 762M) and \textbf{LLaMA} \cite{DBLP:journals/corr/abs-2407-21783} (1B, 3B, 8B).

Following \cite{li2025multihopquestiongenerationdualperspective}, we evaluate the proposed framework using standard automatic metrics: \textbf{BLEU-4} \cite{DBLP:conf/acl/PapineniRWZ02}, \textbf{METEOR} \cite{banerjee-lavie-2005-meteor}, \textbf{ROUGE-L} \cite{lin-2004-rouge}, and \textbf{BERTScore} \cite{DBLP:conf/iclr/ZhangKWWA20}.

Due to space limitations, detailed descriptions of the baselines and evaluation metrics are provided in the supplementary material.

\paragraph{Implementation Details}
We optimize the models using AdamW, incorporating learning rate warm-up and gradient clipping. For the HotpotQA dataset, the (batch size, learning rate) configurations are $(16, 5 \times 10^{-5})$, $(4, 2.5 \times 10^{-5})$, and $(4, 5 \times 10^{-6})$ for the GPT (117M, 345M, 762M) variants, respectively; for LLaMA (1B, 3B, 8B), they are $(4, 1 \times 10^{-4})$, $(4, 1 \times 10^{-5})$, and $(2, 5 \times 10^{-5})$. On the MuSiQue dataset, the corresponding LLaMA configurations are $(2, 1 \times 10^{-5})$, $(2, 5 \times 10^{-5})$, and $(2, 1 \times 10^{-4})$. Here, we jointly train across the 2-, 3-, and 4-hop instances, while evaluating each subset independently. All LLaMA models are fine-tuned using LoRA \cite{hulora} ($r=8$, $\alpha=32$). Following standard contrastive learning practice~\cite{he2020momentum}, we set the temperature parameter to $\tau_2 = 0.07$. We also set $\tau_1 = 0.07$. To ensure a fair comparison, all backbones follow the same experimental setup and are trained only with the MQG and QA learning objectives.

\subsection{Results and Analysis}
\label{results_and_analysis}
Table~\ref{mainresult_merged} compares the proposed QQ framework with advanced baselines on the HotpotQA and MuSiQue datasets. The results show that QQ consistently improves different backbones across multiple metrics. In particular, LLaMA$_{\text{8B}}$ w/ QQ achieves the best results on most evaluation metrics, demonstrating the superiority of the framework. The advantages of QQ are observed across different levels of task difficulty. On HotpotQA, smaller backbones, such as the GPT series, obtain larger improvements in the SF setting, whereas larger backbones, such as the LLaMA series, benefit more in the Full setting. This suggests that context length poses different challenges to architectures of different scales. On the more challenging MuSiQue dataset, which requires more complex multi-hop reasoning, QQ consistently yields substantial performance gains, especially in the 2-hop and 4-hop settings. These results indicate that the proposed bidirectional alignment mechanism effectively facilitates utility transfer between MQG and QA, particularly in scenarios with higher reasoning complexity.

Further analysis shows that overall generation performance generally improves as the number of parameters increases, while QQ consistently narrows the gap between vanilla backbones and high-quality question generation. On MuSiQue, although BLEU-4 shows an unusual peak in the 3-hop setting, possibly due to a favorable context length under this configuration, METEOR and ROUGE-L generally decrease as reasoning complexity increases, which aligns with the expected increase in task difficulty. In addition, variants such as LLaMA$_{\text{(1B)}}$ achieve substantial relative improvements, but their absolute performance remains limited by the capacity of the backbone itself. Notably, QQ integrates MQG and QA within a unified framework, facilitating the construction of more versatile agents. Despite slight fluctuations in individual metrics, QQ consistently outperforms the vanilla backbones, demonstrating the stability and robustness of leveraging the inherent duality between MQG and QA to improve generation quality.

\begin{table}[!t]
	\centering
		\begin{tabular}{lccc}
			\toprule
			Model          & Fluency & Relevance & Complexity \\ \midrule
			DPKG           & 4.74    & 4.69      & 3.57       \\
			LLaMA$_{\text{(8B)}}$        & 4.65    & 4.73      & 3.49       \\
			LLaMA$_{\text{(8B)}}$   w/ QQ & 4.75    & 4.84      & 3.72       \\
			Ground-truth   & 4.80    & 4.85      & 4.29       \\ \midrule
			kappa          & 0.81    & 0.75      & 0.66       \\ \bottomrule
	\end{tabular}
	\caption{Pointwise human evaluation results. Fleiss' kappa is used to measure inter-annotator agreement.}
	\label{human_pointwise}
\end{table}

\begin{table*}[!t]
	\centering
	\small
	\renewcommand{\arraystretch}{1.3}
	\begin{tabularx}{\textwidth}{@{} >{}p{3.2cm} X >{\raggedright\arraybackslash}p{3.5cm} @{}}
		\toprule
		\textnormal{} & Question & Answer \\
		\midrule
		
		Documents &
		\multicolumn{2}{p{\dimexpr\textwidth-3.2cm-2\tabcolsep\relax}}{%
			\textit{(Actaea (plant))} Actaea, commonly called baneberry, bugbane and cohosh, is a genus of flowering plants belonging to the family Ranunculaceae, native to subtropical, temperate and subartic regions of the Europe, Asia and North America. \par \vspace{0.4em}
			
			\textit{(Onoclea)} Onoclea is a genus of plants in the Onocleaceae family, native to moist habitats in eastern Asia and eastern North America. They are deciduous ferns with sterile fronds arising from creeping rhizomes in spring, dying down at first frost. Fertile fronds appear in late summer. Depending on the authority, the genus contains one to five species.
		} \\
		\midrule
		
		LLaMA$_{\text{(8B)}}$ &
		What continent are both Actaea and Onoclea found? &
		Asia \\
		
		LLaMA$_{\text{(8B)}}$ w/ QQ &
		What continent is Actaea native to that Onoclea is not? &
		Europe \\
		\midrule
		
		Ground Truth &
		The  Actaea genus is native to what continent that the Onoclea genus is not? &
		Europe \\
		\bottomrule
	\end{tabularx}
	\caption{Example of question and answer generation using LLaMA$_{\text{(8B)}}$ with vs. without the QQ framework.}
	\label{case_table_}
\end{table*}

\begin{figure*}[!t]
	\centering
	\includegraphics[width=\textwidth]{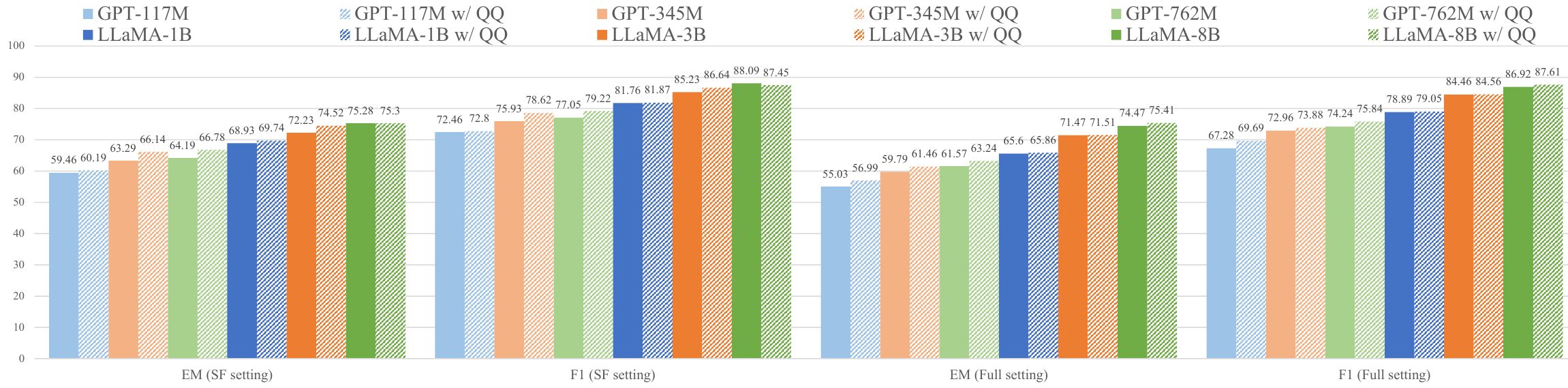}
	\caption{Results of question answering with generated questions on the HotpotQA dataset.
	}
	\label{qa_perf}
\end{figure*}

\subsection{Answerability of Generated Questions}
\label{answer_of_questions}
A key feature of the proposed framework is its joint support for multi-hop question generation and question answering, a capability that remains largely unexplored in prior work. We leverage this capability to evaluate the answerability of generated questions using Exact Match (EM) and F1-score (F1), as shown in Figure~\ref{qa_perf}. The results indicate that QA performance generally scales with parameter size, and QQ further improves both EM and F1 across all evaluated architectures. Notably, GPT$_{\text{(345M)}}$ w/ QQ occasionally outperforms larger counterparts, suggesting that our framework can simultaneously enhance both question generation and answer prediction quality. Consistent with the results in Table~\ref{mainresult_merged}, higher MQG evaluation scores generally correspond to improved EM and F1, confirming the enhanced answerability of the generated questions. Although LLaMA$_{\text{(8B)}}$ shows minor discrepancies in QA metrics despite larger gaps in MQG performance, this may be attributed to its stronger semantic understanding ability, which can compensate for minor structural irregularities in generated questions. Overall, these findings validate that the proposed bidirectional alignment between MQG and QA effectively improves both the answerability and structural quality of generated questions.

\subsection{Ablation Study}
We conduct an ablation study on the GPT series backbones to evaluate the effectiveness of each component in the QQ framework, as reported in Table~\ref{ablation_study}. The proposed QQ framework substantially improves the performance of the GPT series, with more pronounced gains in the SF setting than in the Full setting. The alignment loss consistently improves all metrics, whereas the contrastive loss occasionally reduces specific metrics, such as ROUGE-L for GPT$_{\text{(345M)}}$ in the Full setting. This indicates that contrastive loss does not uniformly improve performance, while alignment loss provides a more reliable and stable signal for enhancing question generation. We further observe that the gains from the alignment loss are evident in both the SF and Full settings, whereas the gains from the contrastive loss are slightly larger in the SF setting. Across all backbones, the alignment loss contributes more substantially to performance improvements, particularly for GPT$_{\text{(762M)}}$. Moreover, its impact becomes increasingly pronounced as the parameter scale grows. These results support our view that $\mathcal{L}_{Q \to A}$ and $\mathcal{L}_{A \to Q}$ mutually align MQG-generated questions with their corresponding QA-generated answers, thereby improving the quality of generated questions, while contrastive learning further reinforces this alignment. Overall, these components complement each other, and the QQ framework demonstrates strong potential for improving MQG performance. This suggests that fully exploiting the intrinsic duality between MQG and QA can substantially improve question quality.

\subsection{Human Evaluation}
We conduct human evaluation using LLaMA$_{\text{(8B)}}$ as the backbone and compare it with DPKG$_{\text{hard}}$, an advanced baseline, on 300 randomly sampled questions from the HotpotQA Full setting. Three graduate students evaluate the generated questions along three dimensions: (i) \textit{Fluency}, which measures whether the question is grammatically correct and logically coherent; (ii) \textit{Relevance}, which evaluates whether the question is answerable and contextually appropriate; and (iii) \textit{Complexity}, which assesses whether the question requires reasoning over multiple document snippets. Further details are provided in the supplementary material. 
Table~\ref{human_pointwise} presents the evaluation results. The inter-annotator agreement, measured by Fleiss' kappa, ranges from 0.66 to 0.81, indicating high reliability. The relatively lower agreement on \textit{Complexity} compared to \textit{Fluency} and \textit{Relevance} suggests that human judgment on this dimension is slightly more inconsistent, yet it remains reliable. For \textit{Relevance}, LLaMA$_{\text{(8B)}}$ w/ QQ achieves a score of 4.84, closely approaching the ground truth and substantially outperforming all baselines. This confirms that QQ effectively grounds the generation process in the provided context. For \textit{Complexity}, LLaMA$_{\text{(8B)}}$ w/ QQ notably improves over both the standard LLaMA and DPKG, suggesting that exploiting the intrinsic duality between MQG and QA encourages deeper logical reasoning. For \textit{Fluency}, all evaluated variants achieve high scores, while LLaMA$_{\text{(8B)}}$ w/ QQ maintains a slight advantage, indicating that the backbone's inherent grammatical capability is well preserved. Overall, these human evaluation results are consistent with the automatic evaluation results, validating the practical effectiveness of QQ in multi-hop question generation.

\subsection{Case Study}
To further illustrate the semantic quality of generated questions and highlight the contribution of the QQ framework, we conduct a case study comparing LLaMA$_{\text{(8B)}}$ with and without QQ. The case is presented in Table~\ref{case_table}, where the answer is generated conditioned on the question produced by the unified model, i.e., the answer aligned with the generated question. It is evident that the answer generated by LLaMA$_{\text{(8B)}}$ w/ QQ is factually correct and more faithful to the ground truth than that produced by LLaMA$_{\text{(8B)}}$. Moreover, the question generated by LLaMA$_{\text{(8B)}}$ w/ QQ is of substantially higher quality: it preserves the key comparative nuance of the ground truth, whereas LLaMA$_{\text{(8B)}}$ alters the intended reasoning relation into a fundamentally different question. Specifically, the ground truth asks which continent \textit{Actaea} is native to that \textit{Onoclea} is not, requiring the model to identify a set-difference relation across the two documents. The question generated by LLaMA$_{\text{(8B)}}$ w/ QQ successfully retains the crucial phrase, \textit{that Onoclea is not}, which is semantically aligned with the ground-truth expression \textit{that the Onoclea genus is not}. In contrast, LLaMA$_{\text{(8B)}}$ asks which continent both \textit{Actaea} and \textit{Onoclea} are found in, thereby converting the original difference-based reasoning into an intersection-based query. Although the question generated by LLaMA$_{\text{(8B)}}$ w/ QQ is slightly more concise than the ground truth, it remains highly comparable in semantic richness and substantially outperforms LLaMA$_{\text{(8B)}}$ in preserving the intended multi-document reasoning.
Overall, this case study demonstrates that QQ improves both the linguistic and semantic quality of generated questions, further confirming the effectiveness of aligning MQG-generated questions with their corresponding QA-generated answers.


\section{Conclusion}
This paper proposes QQ, a unified framework that supports both MQG and QA. QQ explicitly models the intrinsic duality between the two tasks through bidirectional alignment constraints, thereby improving the quality of generated multi-hop questions, as validated by extensive evaluations. Despite its effectiveness, this study has several limitations. Due to hardware constraints, we mainly evaluate QQ on GPT- and LLaMA-series backbones. Future work will incorporate context-consistency constraints to strengthen the alignment among context modeling, question generation, and answer generation. In addition, since a single answer may correspond to multiple valid questions, our predictability-based alignment loss defines a broader positive-pair signal than standard contrastive learning, whose theoretical basis remains to be further refined.

\bibliography{AnonymousSubmission2027}

\section{Datasets}
\label{appdix_dataset}
We conduct experiments on two widely used MQG datasets: \textbf{HotpotQA}~\cite{yang-etal-2018-hotpotqa} and \textbf{MuSiQue}~\cite{trivedi-etal-2022-musique}.

HotpotQA is a large-scale Wikipedia-based benchmark designed for multi-hop reasoning over multiple documents, containing approximately 10K crowdsourced question-answer pairs. Each question is associated with two supporting documents that provide the necessary evidence for answer inference. HotpotQA provides two evaluation settings: supporting facts (SF) and full document context (Full). In the SF setting, only sentences containing the supporting facts for the answer are provided. In contrast, the Full setting provides longer and more complex documents, resulting in a more challenging evaluation scenario. Since the original test set is not publicly available, we adopt the data split provided by \cite{li-etal-2025-multi-hop}. Detailed statistics are summarized in Table~\ref{table_dataset}.

MuSiQue contains 2-hop, 3-hop, and 4-hop reasoning questions constructed through a bottom-up procedure, with approximately 25K samples in its Ans setting. It provides two settings: \textbf{Ans} and \textbf{Full}. In the Ans setting, all questions are answerable. In the Full setting, each sample includes both an answerable and an unanswerable question, effectively doubling the dataset size. Since this work focuses on multi-hop question generation, we use only the Ans setting and partition the dataset accordingly. Specifically, we randomly sample a small portion of the original development set as our new development set and reserve the remaining samples for testing. Detailed statistics are reported in Table~\ref{table_dataset_musique}. We will release our partitioned dataset to facilitate future research.

Together, these two datasets provide a robust testbed for evaluating the ability of the proposed framework to exploit the intrinsic duality between MQG and QA through bidirectional alignment, which substantially improves the quality of generated multi-hop questions.

\begin{table}[!t]
	\small
	\centering
	\begin{tabular}{lccc}
		\toprule
		\multirow{2}{*}{} & \multirow{2}{*}{\#Sample} & \multicolumn{2}{c}{Average context length} \\ \cmidrule(l){3-4} 
		&                           & Full setting             & SF setting            \\ \midrule
		Train             & 89947                     & 204.4               & 96.1                 \\
		Dev               & 500                       & 199.8               & 95.5                 \\
		Test              & 7405                      & 205.8               & 98.5                 \\ \bottomrule
	\end{tabular}
	\caption{Statistics of the HotpotQA dataset.}
	\label{table_dataset}
\end{table}

\begin{table}[!t]
	\centering
	\setlength{\tabcolsep}{0.8mm}{
		\begin{tabular}{lccccc}
			\toprule
			& \#2-hop    & \#3-hop    & \#4-hop & \#Total & Len.  \\ \cmidrule(l){2-6} 
			Train    & 14376      & 4387       & 1175    & 19938   & 362.3 \\
			Dev      & 600        & 350        & 200     & 1150    & 368.8 \\
			Test     & 652        & 410        & 205     & 1267    & 369.5 \\ \midrule
			\multicolumn{3}{c}{\#2-hop   Len.} & \multicolumn{3}{c}{238.4} \\
			\multicolumn{3}{c}{\#3-hop   Len.} & \multicolumn{3}{c}{385.9} \\
			\multicolumn{3}{c}{\#4-hop   Len.} & \multicolumn{3}{c}{476.3} \\ \bottomrule
	\end{tabular}}
	\caption{Statistics of the MuSiQue dataset. "Len." denotes "average context length."}
	\label{table_dataset_musique}
\end{table}

\begin{figure}[!t]
	\centering
	\includegraphics[width=0.45\textwidth]{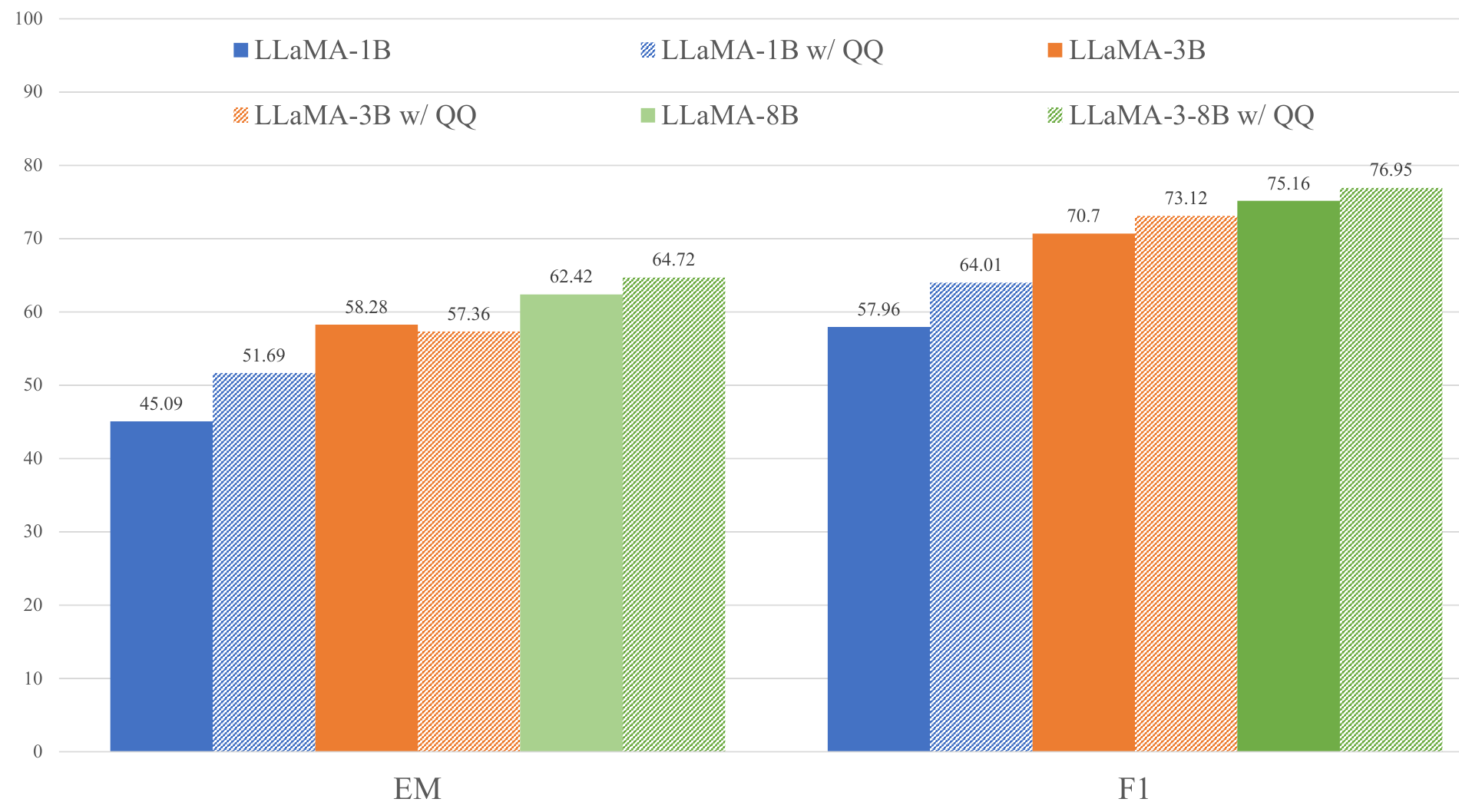}
	\caption{Results of question answering using generated questions on the MuSiQue dataset (2-hop).
	}
	\label{qa_perf2hop}
\end{figure}

\begin{figure}[!t]
	\centering
	\includegraphics[width=0.45\textwidth]{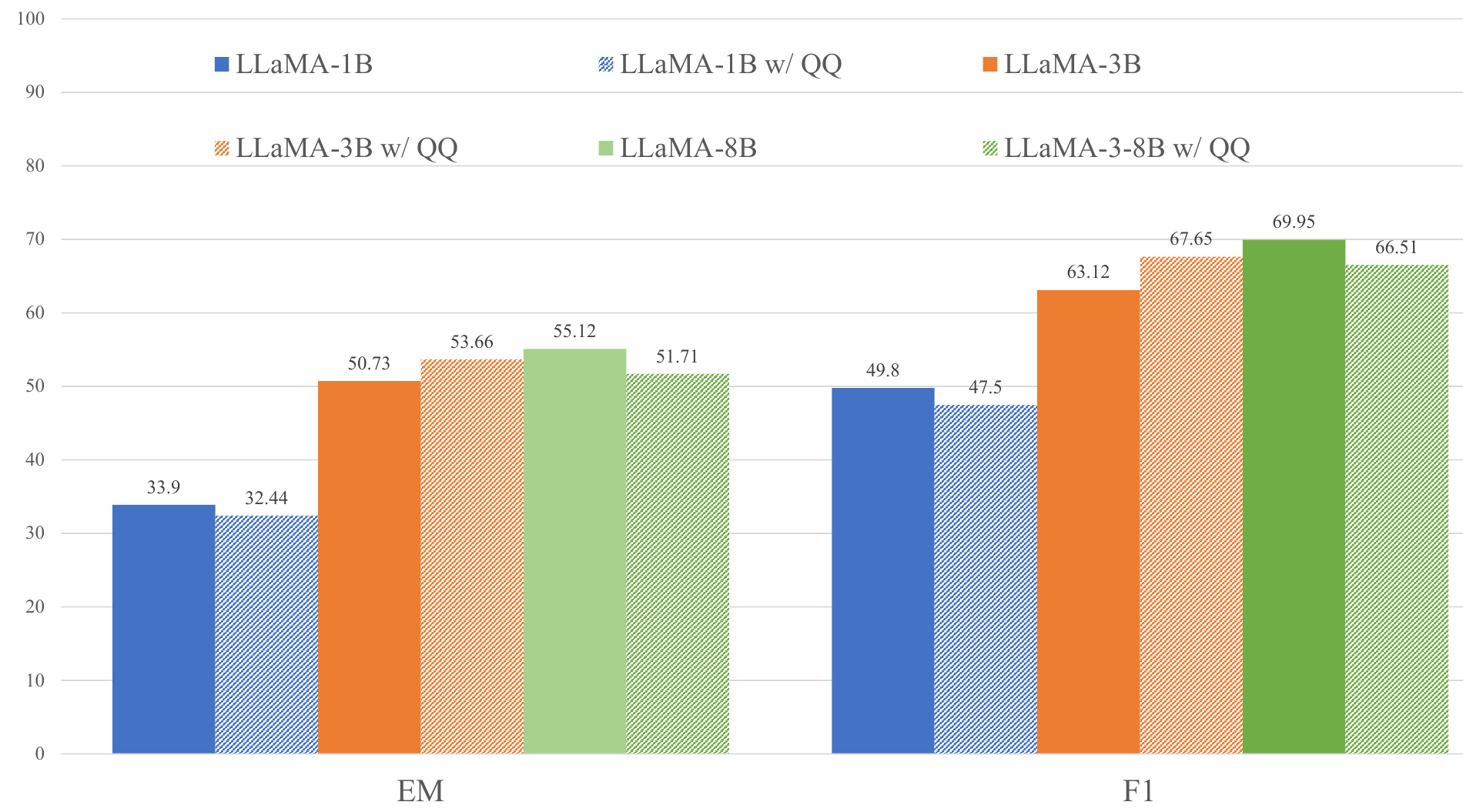}
	\caption{Results of question answering using generated questions on the MuSiQue dataset (3-hop).
	}
	\label{qa_perf3hop}
\end{figure}

\begin{figure}[!t]
	\centering
	\includegraphics[width=0.45\textwidth]{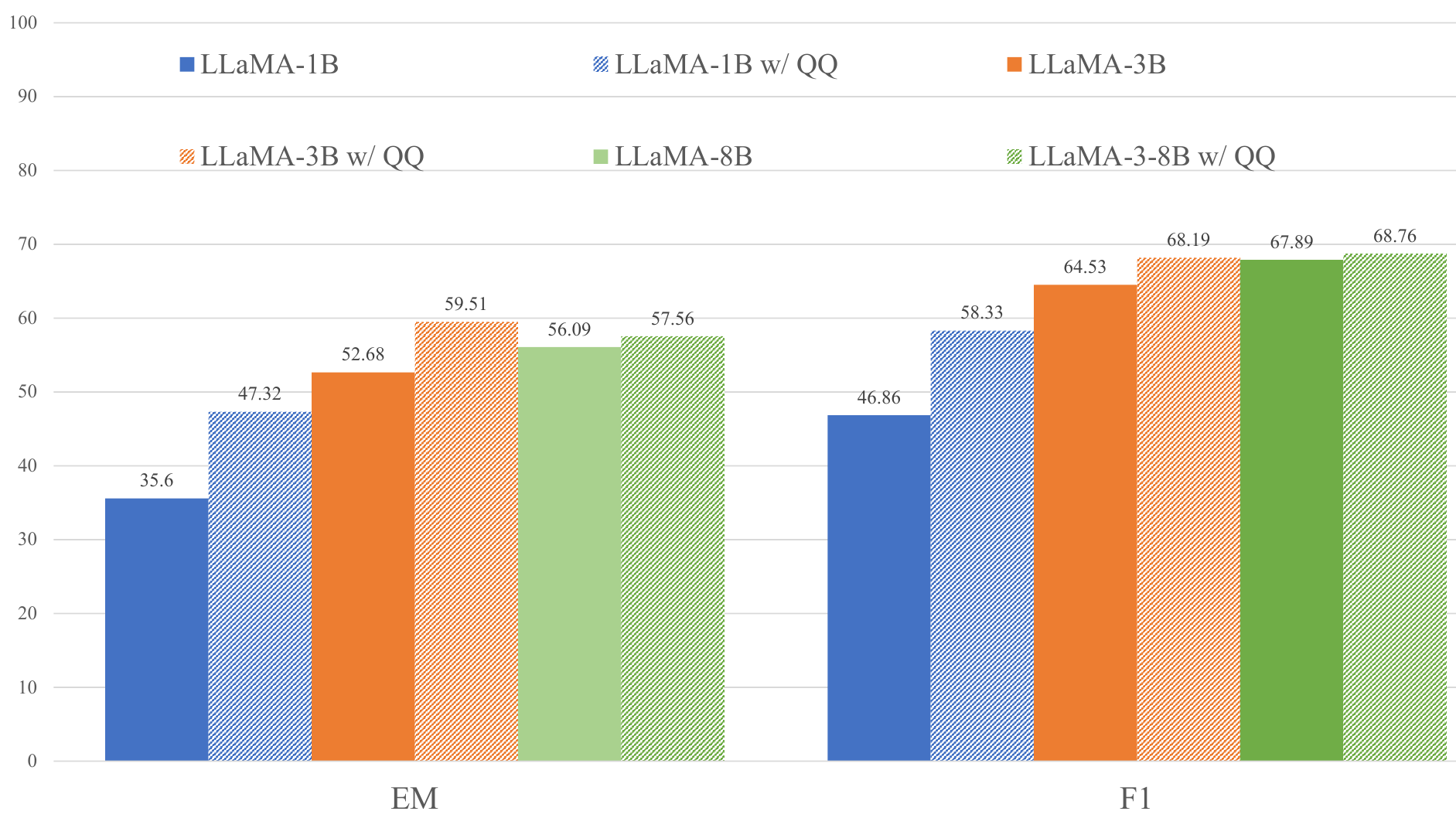}
	\caption{Results of question answering using generated questions on the MuSiQue dataset (4-hop).
	}
	\label{qa_perf4hop}
\end{figure}

\section{Baselines}
\label{baselines_app}
We compare the QQ framework against representative state-of-the-art models. For HotpotQA, baselines include:
\begin{itemize}
	\item \textbf{MulQG} \cite{su-etal-2020-multi}: Incorporates GNNs into multi-hop reasoning.
	\item \textbf{CQG} \cite{fei-etal-2022-cqg}: Controls key entities during decoding to improve generation quality.
	\item \textbf{MixQG} \cite{murakhovska-etal-2022-mixqg}: A generative model trained on diverse QA datasets.
	\item \textbf{QA4QG} \cite{su2022qa4qg}: Leverages an auxiliary QA model to constrain generation.
	\item \textbf{E2EQR} \cite{DBLP:conf/coling/Hwang0L24}: Enhances question complexity via sequential rewriting.
	\item \textbf{SGCM} \cite{ding-etal-2024-sgcm}: Identifies salient document sentences for guidance.
	\item \textbf{DPKG$_{\text{hard/soft}}$} \cite{li-etal-2025-multi-hop}: Utilizes dual-perspective keyword guidance.
\end{itemize}
For MuSiQue, in addition to the aforementioned models, we include \textbf{DP-Graph} \cite{pan-etal-2020-semantic}, which utilizes both word-level and node-level semantic graph representations.

\begin{figure}[!t]
	\centering
	\includegraphics[width=\columnwidth]{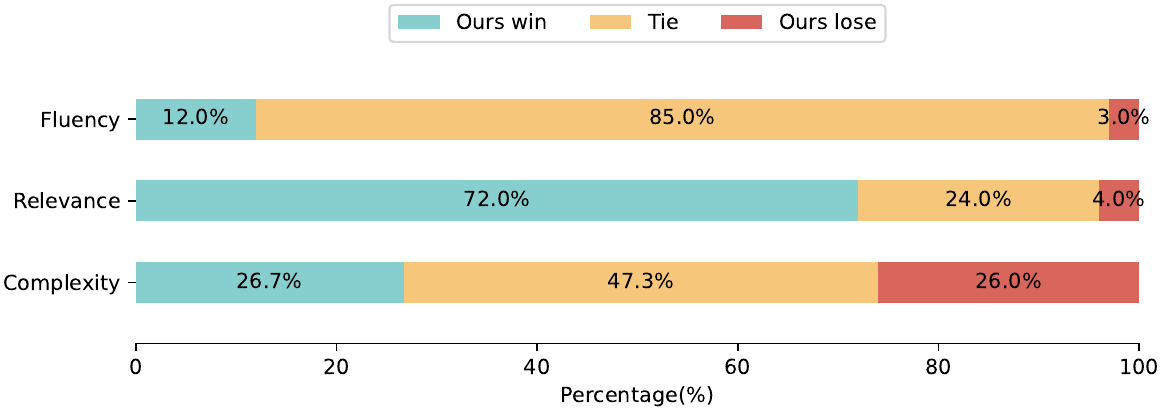}
	\caption{Pairwise human evaluation results.
	}
	\label{figure_human}
\end{figure}

\begin{table*}[!t]
	\centering
	\small
	\renewcommand{\arraystretch}{1.3} 
	\begin{tabularx}{\textwidth}{@{} >{}p{3.2cm} X >{\raggedright\arraybackslash}p{3.5cm} @{}}
		\toprule
		\textnormal{} & Question & Answer \\
		\midrule
		
		Documents &
		\multicolumn{2}{p{\dimexpr\textwidth-3.2cm-2\tabcolsep\relax}}{%
			\textit{(Das Damen)} Das Damen was an alternative rock band from New York City, United States, formed in 1984. The band released several albums before splitting up in 1991. The band's name is fake German and allegedly translates to "the ladies" (the correct German form would be "Die Damen"). \par \vspace{0.4em}
			
			\textit{(Sugar Ray)} Sugar Ray is an American rock band formed in 1986. The band, starting off more as a funk metal band, gained mainstream fame in 1997 with their release of the song "Fly". This song's success, coupled with its pop rock sound that was quite different from the rest of their material at the time, led the band to change to a mainstream, pop music style. Subsequent albums shared this style, and the band landed a number of hits with "Every Morning" and "Someday" and "When It's Over".
		} \\
		\midrule
		
		LLaMA$_{\text{(3B)}}$ &
		What kind of rock did Das Damen and Sugar Ray play? &
		rock \\
		
		LLaMA$_{\text{(3B)}}$ w/ QQ &
		What type of rock music were both Das Damen and Sugar Ray known for? &
		alternative rock \\
		\midrule
		
		Ground Truth &
		The rock band Sugar Ray began as a funk metal band, while the band Das Damen was considered what? &
		alternative rock \\
		\bottomrule
	\end{tabularx}
	\caption{Example of question and answer generation using LLaMA$_{\text{(3B)}}$ (with vs. without the QQ framework).}
	\label{case_table}
\end{table*}

\begin{table*}[!t]
	\centering
	\setlength{\tabcolsep}{1.0mm}{
		\begin{tabular}{lcccccc}
			\toprule
			\multirow{2}{*}{Model} & \multicolumn{3}{c}{SF   setting} & \multicolumn{3}{c}{Full   setting} \\ \cmidrule(l){2-7} 
			& BLEU-4   & METEOR  & ROUGE-L  & BLEU-4   & METEOR   & ROUGE-L   \\ \midrule
			Qwen-plus   w/ zero-shot       & 13.60    & 19.56   & 33.88    & 10.62    & 16.82    & 30.34     \\
			Qwen-plus   w/ one-shot       & 14.97    & 20.68   & 34.42    & 11.17    & 17.37    & 30.39     \\ \cmidrule(l){2-7} 
			Qwen-plus$_{gt}$   w/ zero-shot       & 20.35    & 25.44   & 44.28    & 18.77    & 24.14    & 42.71     \\
			Qwen-plus$_{gt}$   w/ one-shot       & 22.81    & \underline{27.57}   & 46.46    & 20.91    & \textbf{26.18}    & \textbf{44.88}     \\ \cmidrule(l){2-7} 
			LLaMA-3$_{\text{(8B)}}$                & \underline{25.90}    & 27.30   & \textbf{47.12}    & \underline{21.18}    & 23.44    & 43.66     \\
			LLaMA-3   w/ QQ        & \textbf{26.83}    & \textbf{27.96}   & \underline{46.84}    & \textbf{23.71}    & \underline{25.41}    & \underline{44.20}     \\ \bottomrule
	\end{tabular}}
	\caption{Performance comparison between our framework and prompted LLMs on multi-hop question generation. \textbf{Bold} text highlights the best results, while \underline{underlined} text indicates the second-best. Here, w/ denotes with.}
	\label{prompt_llm}
\end{table*}

\begin{figure*}[!t]
	\centering
	\includegraphics[width=\textwidth]{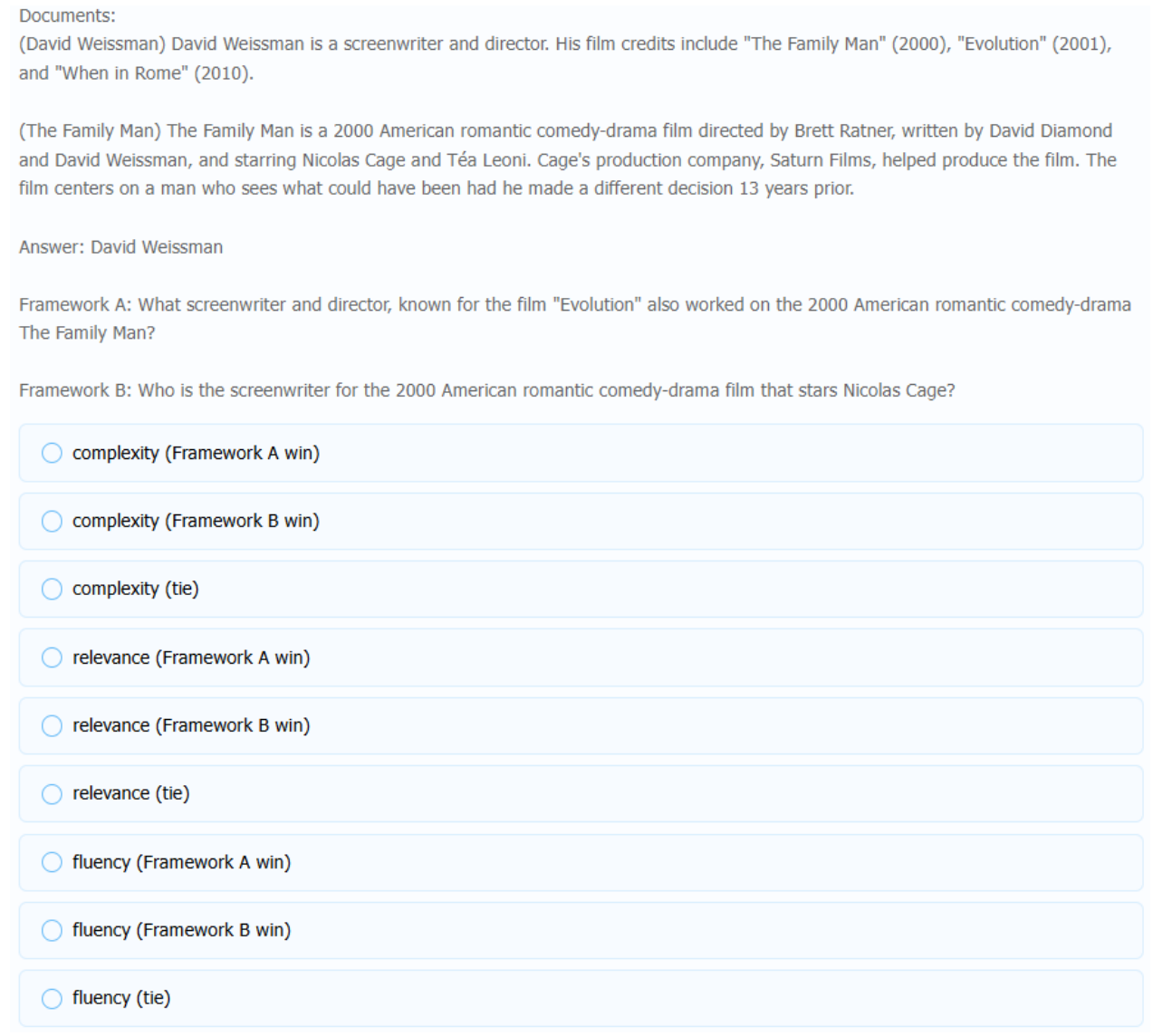}
	\caption{Human evaluation form (pairwise human evaluation). Framework A refers to LLaMA$_{\text{(3B)}}$ w/ QQ, and Framework B refers to LLaMA$_{\text{(3B)}}$. During evaluation, the evaluators are blinded to the correspondence between the frameworks and the models they represent.}
	\label{human_eva}
\end{figure*}

\begin{figure*}[!t]
	\centering
	\includegraphics[width=\textwidth]{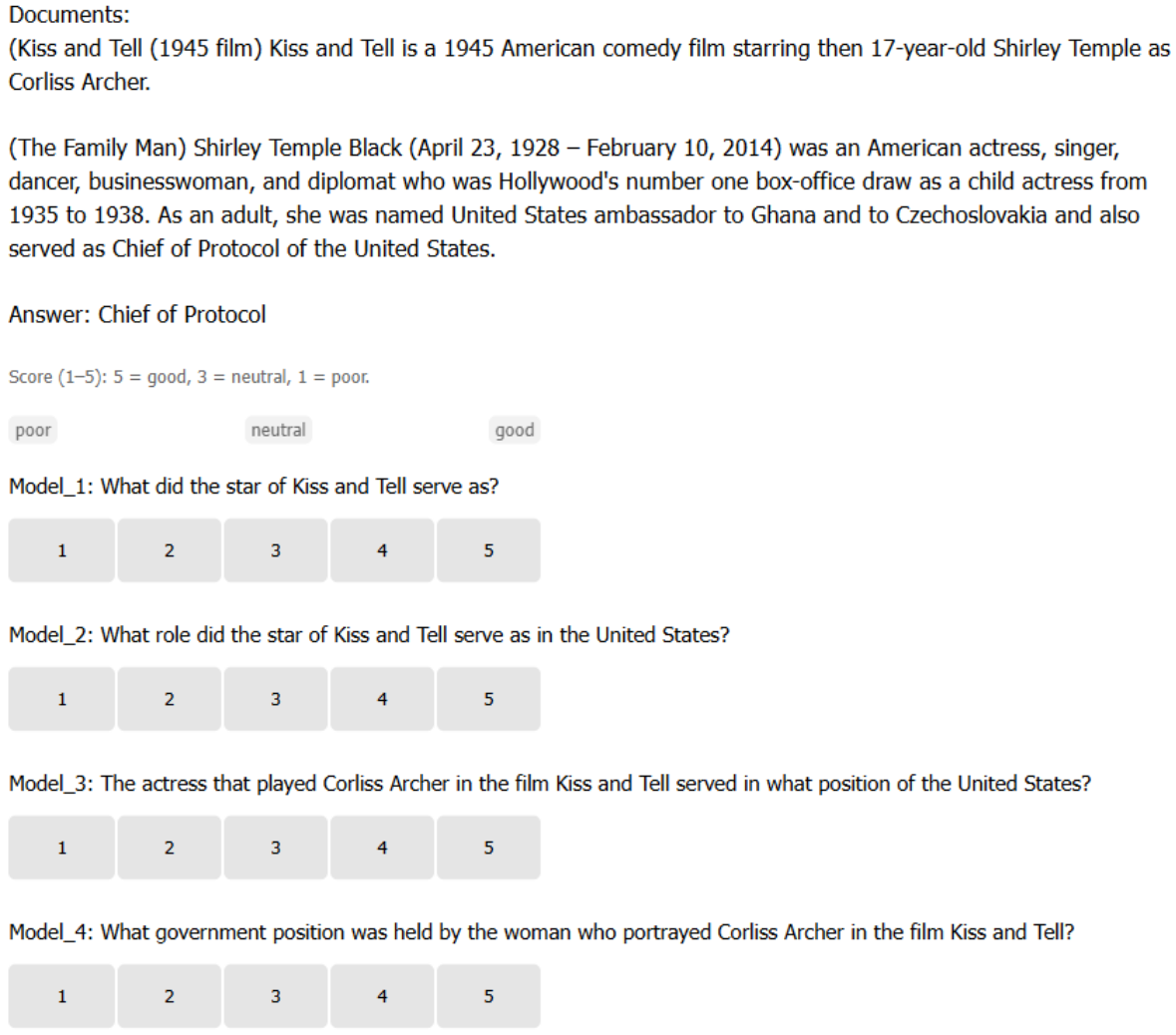}
	\caption{Human evaluation form (pointwise human evaluation).  Model\_1, Model\_2, Model\_3, and Model\_4 denote DPKG, LLaMA$_{\text{(8B)}}$, LLaMA$_{\text{(8B)}}$ w/ QQ, and the ground-truth question, respectively. During evaluation, the evaluators are blinded to the correspondence between the model identifiers and the models they represent.
	}
	\label{human_eva2}
\end{figure*}

\section{Evaluation Metrics}
\label{evaluation_metrics_app}
Following prior work \cite{ding-etal-2024-sgcm,li-etal-2025-multi-hop}, we employ the following automatic metrics: 
\begin{itemize}
	\item \textbf{BLEU-4} \cite{DBLP:conf/acl/PapineniRWZ02} for lexical precision;
	\item \textbf{METEOR} \cite{banerjee-lavie-2005-meteor} for semantic alignment via stemming and synonym matching;
	\item \textbf{ROUGE-L} \cite{lin-2004-rouge} for structural and fluency evaluation;
	\item \textbf{BERTScore} \cite{DBLP:conf/iclr/ZhangKWWA20} for contextual semantic similarity.
\end{itemize}
Additionally, to assess the QA capability of our unified framework, we report \textbf{Exact Match (EM)} and \textbf{F1-score (F1)} for answerability evaluation.

\section{Answerability of Generated Questions (MuSiQue)}
\label{app_musique}
We leverage the QA capability of our unified framework to assess the answerability of generated questions. The results for the 2-hop, 3-hop, and 4-hop settings are presented in Figures~\ref{qa_perf2hop}, \ref{qa_perf3hop}, and \ref{qa_perf4hop}, respectively.

In the 2-hop and 4-hop settings, EM and F1 generally scale with parameter size, and QQ consistently improves performance over the vanilla backbones. In the 3-hop setting, this trend remains consistent for the 3B variants, while slight fluctuations are observed for the 1B and 8B architectures with QQ. These variations are expected, as automatic MQG metrics and QA-based answerability metrics capture different aspects of question quality. Strong QA models may still answer correctly even when the generated questions contain minor formulation issues. Nevertheless, in most cases, QQ outperforms the vanilla backbones, indicating that answerability evaluation provides complementary evidence for assessing overall question quality. Although some trends on MuSiQue differ slightly from those on HotpotQA, the discrepancies are minor. Overall, these findings demonstrate that QQ improves question generation quality from a QA performance perspective, further confirming the benefit of aligning MQG-generated questions with their corresponding QA-generated answers.

\section{Human Evaluation}
\label{appdix_human_evaluation}
We conduct both pairwise and pointwise human evaluations. The corresponding evaluation forms are shown in Figures~\ref{human_eva} and~\ref{human_eva2}, respectively. While the pointwise evaluation results are reported in the main text, this section focuses on the pairwise evaluation. Specifically, we compare LLaMA$_{\text{(3B)}}$ with and without the proposed QQ framework. We randomly sample 300 questions generated in the Full setting, and the results are presented in Figure~\ref{figure_human}. The labels "win," "tie," and "lose" indicate whether LLaMA$_{\text{(3B)}}$ w/ QQ performs better than, comparably to, or worse than the vanilla backbone, respectively. The evaluation considers three dimensions: (i) \textit{Fluency}, which measures whether the question is grammatically correct and logically coherent; (ii) \textit{Relevance}, which evaluates whether the question is answerable and contextually appropriate; and (iii) \textit{Complexity}, which assesses whether the question requires reasoning over multiple document snippets.

For \textit{Relevance}, LLaMA$_{\text{(3B)}}$ w/ QQ achieves clear advantages, consistent with our observations in the main results: incorporating the QA process into MQG effectively improves answerability and increases the information density of generated questions. For \textit{Fluency}, both variants perform similarly, with QQ showing a slight advantage. For \textit{Complexity}, the two variants remain comparable, while QQ again yields a modest improvement. These results suggest that the backbone, due to its substantial parameter scale, already possesses strong grammatical and reasoning capabilities. Meanwhile, the proposed framework preserves these strengths while further improving overall question generation quality. In summary, the human evaluation results are consistent with the automatic metrics, further validating the practical effectiveness of the QQ framework.

\section{Case Study}
We also conduct a case study using LLaMA$_{\text{(3B)}}$ as the backbone for further analysis, as shown in Table~\ref{case_table_}. It is evident that the answer generated by LLaMA$_{\text{(3B)}}$ w/ QQ is correct and more complete than that produced by LLaMA$_{\text{(3B)}}$. Moreover, the questions generated by LLaMA$_{\text{(3B)}}$ w/ QQ are of higher quality: they more faithfully capture the full meaning of the ground truth, whereas LLaMA$_{\text{(3B)}}$ generates more simplified questions. For example, the question generated by LLaMA$_{\text{(3B)}}$ w/ QQ includes the phrase \textit{known for}, which is semantically similar to the ground-truth expression \textit{was considered what}; this semantic correspondence is absent from the question generated by LLaMA$_{\text{(3B)}}$. Although the question generated by LLaMA$_{\text{(3B)}}$ w/ QQ differs in length from the ground truth, it remains comparable in semantic richness and substantially outperforms LLaMA$_{\text{(3B)}}$, even though questions from both models require reasoning over multiple documents. Overall, this analysis demonstrates that QQ improves the quality of generated questions, further confirming the effectiveness of aligning MQG-generated questions with their corresponding QA-generated answers to enhance question generation performance.


\section{Prompting Large Language Models}
\label{prompt_llm_li}
MQG is inherently more challenging than QA, as it requires deeper semantic understanding and multi-hop reasoning. Despite the strong capabilities of large language backbones, they often suffer from hallucinations in generative multi-hop tasks, leading to poor performance under zero-shot or few-shot prompting~\cite{ushio-etal-2023-practical, liu2024syntheticcontextgenerationquestion}.

To evaluate this issue, we compare the proposed framework with prompting-based large language models, using the dual-perspective keywords proposed by \cite{li2025multihopquestiongenerationdualperspective} to guide generation. As shown in Table~\ref{prompt_llm}, prompting-based methods, such as Qwen-plus, achieve substantially lower performance without keyword guidance. Even when provided with ground-truth keywords, i.e., Qwen-plus$_{\text{gt}}$, these prompted models are generally inferior to our fine-tuned LLaMA$_{\text{(8B)}}$ w/ QQ, except for marginal gains on specific metrics in the Full setting. These findings are consistent with prior studies~\cite{li2025multihopquestiongenerationdualperspective} and suggest that training dedicated backbones with the proposed bidirectional alignment remains more effective than directly prompting large language models. This further highlights the importance of supervised, strategy-guided fine-tuning for complex MQG tasks.


\end{document}